\documentclass{arxiv}
\usepackage[T1]{fontenc}
\usepackage{dfadobe}  
\usepackage{subfiles} 
\usepackage{cite}  
\usepackage{kotex}
\BibtexOrBiblatex
\electronicVersion
\PrintedOrElectronic

\ifpdf \usepackage[pdftex]{graphicx} \pdfcompresslevel=9
\else \usepackage[dvips]{graphicx} \fi

\title[SPnet]%
      {SPnet: Estimating Garment Sewing Patterns from a Single Image}

\author[Seungchan Lim, Sumin Kim, Sung-Hee Lee]
{\parbox{\textwidth}{ \centering Seungchan Lim$^{1}$, Sumin Kim$^{1}$, Sung-Hee Lee$^{1}$
        }
        \\
{\parbox{\textwidth}{\centering $^1$Korea Advanced Institute of Science and Technology, Republic of Korea%
       }
}
}

\begin{document}

 \teaser{
  \includegraphics[width=\linewidth]{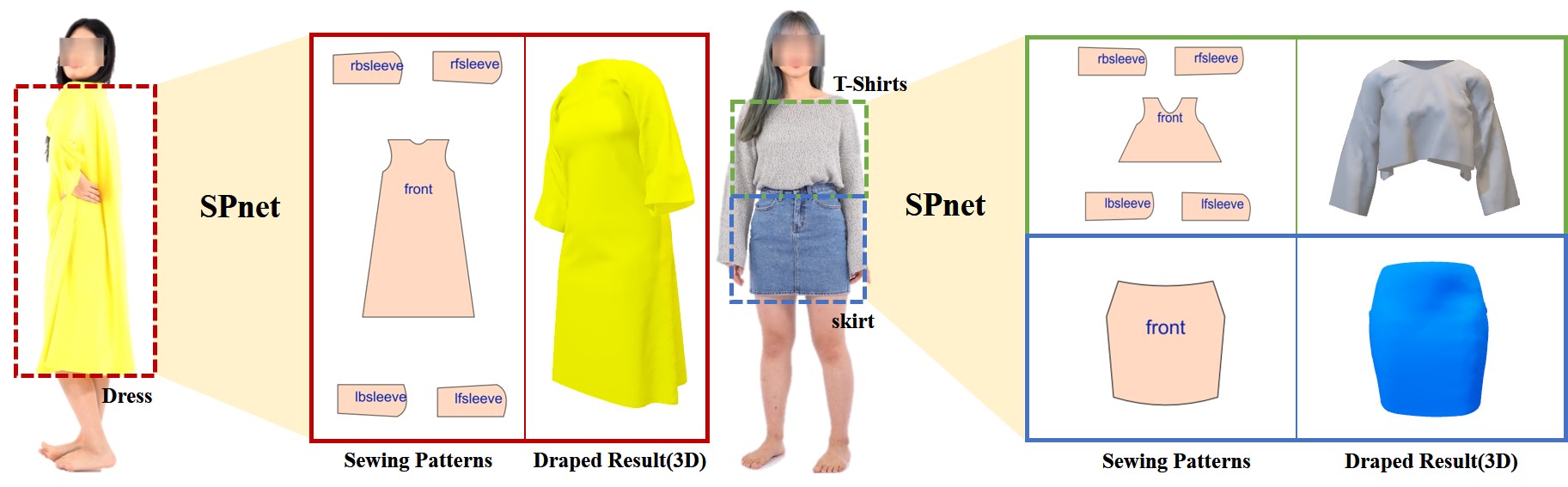}
  \centering
   \caption{\textbf{SPnet} can predict a garment sewing pattern from a single image with an arbitrary pose, enabling the creation of a natural-looking 3D garment mesh. The SPnet is applicable not only to tight-fitting garments but also to loose-fitting garments that exhibit large displacements from the body.}
 \label{fig:teaser}
}

\maketitle
\begin{abstract}
This paper presents a novel method for reconstructing 3D garment models from a single image of a posed user. Previous studies that have primarily focused on accurately reconstructing garment geometries to match the input garment image may often result in unnatural-looking garments when deformed for new poses. To overcome this limitation, our approach takes a different approach by inferring the fundamental shape of the garment through sewing patterns from a single image, rather than directly reconstructing 3D garments. 
Our method consists of two stages. Firstly, given a single image of a posed user, it predicts the garment image worn on a T-pose, representing the baseline form of the garment. Then, it estimates the sewing pattern parameters based on the T-pose garment image. By simulating the stitching and draping of the sewing pattern using physics simulation, we can generate 3D garments that can adaptively deform to arbitrary poses. The effectiveness of our method is validated through ablation studies on the major components and a comparison with other approaches.


\end{abstract}  
\section{Introduction}
\documentclass[../eg2024main.tex]{subfiles}
\begin{document}
%
%
As the reconstruction of the user's clothing into their avatar has gained increasing attention in computer graphics, researchers have developed various techniques to increase accuracy, interactivity, and speed of garment reconstruction.
Regarding input data, various modalities have been investigated, such as 3D scans \cite{NEURIPS2021_eb160de1}, single or multiple images \cite{zhu2020deep}, or video \cite{qiu2023rec}.
Among them, the approach of using a single image \cite{zhu2020deep} has advantages in terms of acquisition cost and accessibility. 
Recent studies \cite{bhatnagar2019multi,jiang2020bcnet,zhu2022registering} show impressive results in reconstructing 3D garment from images.

The range of expressible garments and the naturalness of their shape on a reposed body are significantly influenced by how the garments are represented and deformed. While tight-fitting garments can be efficiently represented as displacements from the skin \cite{bhatnagar2019multi}, loose garments such as skirts and dresses require a separate modeling approach independent from the body. One effective method to achieve this is by utilizing a 3D parametric template to represent the garments \cite{jiang2020bcnet,corona2021smplicit}. However, when the template model is deformed using the linear blend skinning (LBS) method, which constrains garment deformation to a linear relation with the body pose, the quality of the deformation can be significantly compromised \cite{bhatnagar2019multi,jiang2020bcnet,corona2021smplicit}.  


To address these issues, we propose a novel approach to reconstructing garment from a single image. 
Instead of directly reconstructing the 3D garment, our deep learning framework, named \emph{SPnet}, predicts the \textbf{s}ewing \textbf{p}attern of the garment. Sewing patterns are not only informative for defining the basic shape of a garment but also enable the exploration of shape variations based on user poses through cloth simulation techniques. 

However, predicting sewing pattern from images with arbitrary poses poses a significant challenge due to severe occlusion and deformation, including wrinkles, in the garment. To solve this problem, we develop a two-stage deep learning framework. In the first stage, our T-pose garment predictor converts the input image with an arbitrary pose into the shape of clothes in a T-pose, which best reveals the original shape of the garment without the effect of pose variation. Since the overall shape of the garment and the presence of wrinkles can vary according to the pose, even when the same clothes are worn, the challenge lies in separately extracting and analyzing the pose and garment information from the input image.

In the subsequent stage, we predict the parameters to create garment pattern based on the predicted T-pose garment. These predicted garment parameters are then used to create the garment pattern, which is subsequently stitched and draped onto an avatar using a physics simulator, resulting in the generation of a natural-looking garment. 

We evaluate our novel T-pose garment predictor through a series of ablation studies to verify the effect of major components. Furthermore, we demonstrate the effectiveness of our approach by comparing with related studies. 

The main contributions of our work are as follows. First, we propose a unique approach to predict the fundamental shape of garments by directly inferring garment patterns from images, enabling the generation of natural deformation and wrinkles in response to pose changes. Our work is the first deep-learning approach to predict garment sewing patterns from a single image.
Secondly, we introduce a two-stage network that predicts the appearance of clothes in a T-pose, which represents the basic shape of garments, before predicting the garment pattern parameters. This two-stage approach enables our SPnet to predict garment pattern parameters robustly from images of a dressed user with arbitrary poses.



\newcommand{\etalchar}[1]{$^{#1}$}
\begin{thebibliography}{\uppercase{APMTM19}}

\bibitem[APMTM19]{alldieck2019tex2shape}
\textsc{Alldieck T., Pons-Moll G., Theobalt C., Magnor M.}:
\newblock Tex2shape: Detailed full human body geometry from a single image.
\newblock In \emph{Proceedings of the IEEE/CVF International Conference on Computer Vision} (2019), pp.~2293--2303.

\bibitem[BGK{\etalchar{*}}13]{berthouzoz2013parsing}
\textsc{Berthouzoz F., Garg A., Kaufman D.~M., Grinspun E., Agrawala M.}:
\newblock Parsing sewing patterns into 3d garments.
\newblock \emph{Acm Transactions on Graphics (TOG) 32}, 4 (2013), 1--12.

\bibitem[BKC17]{badrinarayanan2017segnet}
\textsc{Badrinarayanan V., Kendall A., Cipolla R.}:
\newblock Segnet: A deep convolutional encoder-decoder architecture for image segmentation.
\newblock \emph{IEEE transactions on pattern analysis and machine intelligence 39}, 12 (2017), 2481--2495.

\bibitem[BKL21]{bang2021estimating}
\textsc{Bang S., Korosteleva M., Lee S.-H.}:
\newblock Estimating garment patterns from static scan data.
\newblock In \emph{Computer Graphics Forum} (2021), vol.~40, Wiley Online Library, pp.~273--287.

\bibitem[BSBC12]{brouet2012design}
\textsc{Brouet R., Sheffer A., Boissieux L., Cani M.-P.}:
\newblock Design preserving garment transfer.
\newblock \emph{ACM Transactions on Graphics 31}, 4 (2012), Article--No.

\bibitem[BTTPM19]{bhatnagar2019multi}
\textsc{Bhatnagar B.~L., Tiwari G., Theobalt C., Pons-Moll G.}:
\newblock Multi-garment net: Learning to dress 3d people from images.
\newblock In \emph{Proceedings of the IEEE/CVF international conference on computer vision} (2019), pp.~5420--5430.

\bibitem[CPA{\etalchar{*}}21]{corona2021smplicit}
\textsc{Corona E., Pumarola A., Alenya G., Pons-Moll G., Moreno-Noguer F.}:
\newblock Smplicit: Topology-aware generative model for clothed people.
\newblock In \emph{Proceedings of the IEEE/CVF conference on computer vision and pattern recognition} (2021), pp.~11875--11885.

\bibitem[DJW{\etalchar{*}}06]{decaudin2006virtual}
\textsc{Decaudin P., Julius D., Wither J., Boissieux L., Sheffer A., Cani M.-P.}:
\newblock Virtual garments: A fully geometric approach for clothing design.
\newblock In \emph{Computer Graphics Forum} (2006), vol.~25, Wiley Online Library, pp.~625--634.

\bibitem[GNK18]{guler2018densepose}
\textsc{G{\"u}ler R.~A., Neverova N., Kokkinos I.}:
\newblock Densepose: Dense human pose estimation in the wild.
\newblock In \emph{Proceedings of the IEEE conference on computer vision and pattern recognition} (2018), pp.~7297--7306.

\bibitem[HLC{\etalchar{*}}18]{huang2018deep}
\textsc{Huang Z., Li T., Chen W., Zhao Y., Xing J., LeGendre C., Luo L., Ma C., Li H.}:
\newblock Deep volumetric video from very sparse multi-view performance capture.
\newblock In \emph{Proceedings of the European Conference on Computer Vision (ECCV)} (2018), pp.~336--354.

\bibitem[HPCL21]{NEURIPS2021_eb160de1}
\textsc{Hong F., Pan L., Cai Z., Liu Z.}:
\newblock Garment4d: Garment reconstruction from point cloud sequences.
\newblock In \emph{Advances in Neural Information Processing Systems} (2021), Ranzato M., Beygelzimer A., Dauphin Y., Liang P., Vaughan J.~W., (Eds.), vol.~34, Curran Associates, Inc., pp.~27940--27951.
\newblock URL: \url{https://proceedings.neurips.cc/paper_files/paper/2021/file/eb160de1de89d9058fcb0b968dbbbd68-Paper.pdf}.

\bibitem[HXL{\etalchar{*}}20]{huang2020arch}
\textsc{Huang Z., Xu Y., Lassner C., Li H., Tung T.}:
\newblock Arch: Animatable reconstruction of clothed humans.
\newblock In \emph{Proceedings of the IEEE/CVF Conference on Computer Vision and Pattern Recognition} (2020), pp.~3093--3102.

\bibitem[JSP{\etalchar{*}}23]{jinka2023sharp}
\textsc{Jinka S.~S., Srivastava A., Pokhariya C., Sharma A., Narayanan P.}:
\newblock Sharp: Shape-aware reconstruction of people in loose clothing.
\newblock \emph{International Journal of Computer Vision 131}, 4 (2023), 918--937.

\bibitem[JSS{\etalchar{*}}20]{jia2020fashionpedia}
\textsc{Jia M., Shi M., Sirotenko M., Cui Y., Cardie C., Hariharan B., Adam H., Belongie S.}:
\newblock Fashionpedia: Ontology, segmentation, and an attribute localization dataset.
\newblock In \emph{Computer Vision--ECCV 2020: 16th European Conference, Glasgow, UK, August 23--28, 2020, Proceedings, Part I 16} (2020), Springer, pp.~316--332.

\bibitem[JZH{\etalchar{*}}20]{jiang2020bcnet}
\textsc{Jiang B., Zhang J., Hong Y., Luo J., Liu L., Bao H.}:
\newblock Bcnet: Learning body and cloth shape from a single image.
\newblock In \emph{Computer Vision--ECCV 2020: 16th European Conference, Glasgow, UK, August 23--28, 2020, Proceedings, Part XX 16} (2020), Springer, pp.~18--35.

\bibitem[KL21]{korosteleva2021generating}
\textsc{Korosteleva M., Lee S.-H.}:
\newblock Generating datasets of 3d garments with sewing patterns.
\newblock \emph{arXiv preprint arXiv:2109.05633} (2021).

\bibitem[KL22]{korosteleva2022neuraltailor}
\textsc{Korosteleva M., Lee S.-H.}:
\newblock Neuraltailor: reconstructing sewing pattern structures from 3d point clouds of garments.
\newblock \emph{ACM Transactions on Graphics (TOG) 41}, 4 (2022), 1--16.

\bibitem[LMR{\etalchar{*}}15]{loper2015smpl}
\textsc{Loper M., Mahmood N., Romero J., Pons-Moll G., Black M.~J.}:
\newblock Smpl: A skinned multi-person linear model.
\newblock \emph{ACM transactions on graphics (TOG) 34}, 6 (2015), 1--16.

\bibitem[QCZ{\etalchar{*}}23]{qiu2023rec}
\textsc{Qiu L., Chen G., Zhou J., Xu M., Wang J., Han X.}:
\newblock Rec-mv: Reconstructing 3d dynamic cloth from monocular videos.
\newblock In \emph{Proceedings of the IEEE/CVF Conference on Computer Vision and Pattern Recognition} (2023), pp.~4637--4646.

\bibitem[RSJ21]{rong2021frankmocap}
\textsc{Rong Y., Shiratori T., Joo H.}:
\newblock Frankmocap: A monocular 3d whole-body pose estimation system via regression and integration.
\newblock In \emph{Proceedings of the IEEE/CVF International Conference on Computer Vision} (2021), pp.~1749--1759.

\bibitem[SHN{\etalchar{*}}19]{saito2019pifu}
\textsc{Saito S., Huang Z., Natsume R., Morishima S., Kanazawa A., Li H.}:
\newblock Pifu: Pixel-aligned implicit function for high-resolution clothed human digitization.
\newblock In \emph{Proceedings of the IEEE/CVF international conference on computer vision} (2019), pp.~2304--2314.

\bibitem[SSSJ20]{saito2020pifuhd}
\textsc{Saito S., Simon T., Saragih J., Joo H.}:
\newblock Pifuhd: Multi-level pixel-aligned implicit function for high-resolution 3d human digitization.
\newblock In \emph{Proceedings of the IEEE/CVF Conference on Computer Vision and Pattern Recognition} (2020), pp.~84--93.

\bibitem[VCR{\etalchar{*}}18]{varol2018bodynet}
\textsc{Varol G., Ceylan D., Russell B., Yang J., Yumer E., Laptev I., Schmid C.}:
\newblock Bodynet: Volumetric inference of 3d human body shapes.
\newblock In \emph{Proceedings of the European conference on computer vision (ECCV)} (2018), pp.~20--36.

\bibitem[VSGC20]{vidaurre2020fully}
\textsc{Vidaurre R., Santesteban I., Garces E., Casas D.}:
\newblock Fully convolutional graph neural networks for parametric virtual try-on.
\newblock In \emph{Computer Graphics Forum} (2020), vol.~39, Wiley Online Library, pp.~145--156.

\bibitem[WCPM18]{wang2018learning}
\textsc{Wang T.~Y., Ceylan D., Popovic J., Mitra N.~J.}:
\newblock Learning a shared shape space for multimodal garment design.
\newblock \emph{arXiv preprint arXiv:1806.11335} (2018).

\bibitem[YAP{\etalchar{*}}16]{yang2016detailed}
\textsc{Yang S., Ambert T., Pan Z., Wang K., Yu L., Berg T., Lin M.~C.}:
\newblock Detailed garment recovery from a single-view image.
\newblock \emph{arXiv preprint arXiv:1608.01250} (2016).

\bibitem[YLG{\etalchar{*}}21]{yoon2021pose}
\textsc{Yoon J.~S., Liu L., Golyanik V., Sarkar K., Park H.~S., Theobalt C.}:
\newblock Pose-guided human animation from a single image in the wild.
\newblock In \emph{Proceedings of the IEEE/CVF Conference on Computer Vision and Pattern Recognition} (2021), pp.~15039--15048.

\bibitem[YPA{\etalchar{*}}18]{yang2018physics}
\textsc{Yang S., Pan Z., Amert T., Wang K., Yu L., Berg T., Lin M.~C.}:
\newblock Physics-inspired garment recovery from a single-view image.
\newblock \emph{ACM Transactions on Graphics (TOG) 37}, 5 (2018), 1--14.

\bibitem[ZCJ{\etalchar{*}}20]{zhu2020deep}
\textsc{Zhu H., Cao Y., Jin H., Chen W., Du D., Wang Z., Cui S., Han X.}:
\newblock Deep fashion3d: A dataset and benchmark for 3d garment reconstruction from single images.
\newblock In \emph{Computer Vision--ECCV 2020: 16th European Conference, Glasgow, UK, August 23--28, 2020, Proceedings, Part I 16} (2020), Springer, pp.~512--530.

\bibitem[ZQQH22]{zhu2022registering}
\textsc{Zhu H., Qiu L., Qiu Y., Han X.}:
\newblock Registering explicit to implicit: Towards high-fidelity garment mesh reconstruction from single images.
\newblock In \emph{Proceedings of the IEEE/CVF Conference on Computer Vision and Pattern Recognition} (2022), pp.~3845--3854.

\bibitem[ZYLD22]{9321139}
\textsc{Zheng Z., Yu T., Liu Y., Dai Q.}:
\newblock Pamir: Parametric model-conditioned implicit representation for image-based human reconstruction.
\newblock \emph{IEEE Transactions on Pattern Analysis and Machine Intelligence 44}, 6 (2022), 3170--3184.
\newblock \href {https://doi.org/10.1109/TPAMI.2021.3050505} {\path{doi:10.1109/TPAMI.2021.3050505}}.

\bibitem[ZYW{\etalchar{*}}19]{9010852}
\textsc{Zheng Z., Yu T., Wei Y., Dai Q., Liu Y.}:
\newblock Deephuman: 3d human reconstruction from a single image.
\newblock In \emph{2019 IEEE/CVF International Conference on Computer Vision (ICCV)} (2019), pp.~7738--7748.
\newblock \href {https://doi.org/10.1109/ICCV.2019.00783} {\path{doi:10.1109/ICCV.2019.00783}}.

\end{thebibliography}
\end{document}